\documentclass[11pt]{article}

\usepackage[margin=1in]{geometry}
\usepackage{amsmath,amssymb,amsfonts}
\usepackage{graphicx}
\usepackage{booktabs}
\usepackage{hyperref}
\usepackage{cleveref}
\usepackage{natbib}
\usepackage{xcolor}
\usepackage{caption}
\usepackage{subcaption}
\usepackage{microtype}
\usepackage{enumitem}

\graphicspath{{./}}

\newcommand{\R}{\mathbb{R}}
\newcommand{\Z}{\mathbb{Z}}

\newcommand{\norm}[1]{\left\|#1\right\|}

\title{Spectral Edge Dynamics Reveal Functional Modes of Learning}

\author{%
  Yongzhong Xu
  \thanks{abbyxu@gmail.com; code at \url{https://github.com/skydancerosel/grokking-integrability}}
}

\date{}

\begin{document}
\maketitle

\begin{abstract}

We study the geometry of training dynamics during grokking through spectral analysis of weight updates in transformer models trained on modular arithmetic.
We confirm the existence of a spectral edge---a small block of leading directions separating from the bulk---which reliably distinguishes grokking from non-grokking regimes across multiple tasks and seeds.

We then investigate the functional content of these directions.
Standard mechanistic interpretability decompositions---including head attribution, activation-space analysis, and sparse autoencoders---fail to capture the spectral edge, indicating that its structure is not localized in parameter or feature space.
Instead, when viewed as perturbations of the model's input-output function, the leading directions exhibit structured behavior in appropriately chosen bases over the input domain.

For symmetry-aligned tasks such as addition and multiplication, the spectral edge collapses to a single dominant Fourier mode when expressed in the correct group-theoretic basis (additive or multiplicative via discrete log).
For subtraction, the edge spans a small family of Fourier modes, while for the quadratic task $x^2+y^2$, no simple one-dimensional harmonic basis suffices.
Instead, its structure is partially explained by cross-terms of additive and multiplicative features, consistent with its compositional form.

Finally, we show that multitask training with a shared trunk amplifies this compositional structure: the spectral edge of the $x^2+y^2$ head becomes more aligned with additive modes characteristic of the addition circuit, providing evidence of functional reuse across tasks.

These results suggest that spectral edge dynamics identify low-dimensional functional subspaces governing learning, whose representation depends on the algebraic structure of the task.
In particular, simple harmonic structure emerges only when the task admits an appropriate symmetry-adapted basis, while more complex tasks require richer functional descriptions.

\end{abstract}

\section{Introduction}
\label{sec:intro}

Training trajectories of neural networks are highly structured, despite the enormous dimensionality of parameter space.
A growing body of work suggests that optimization dynamics concentrate along a small number of dominant directions, particularly near phase transitions such as grokking \citep{xu2026lowdim, xu2026spectral_edge, li2018measuring}.
However, the nature of these directions remains unclear: are they localized circuits, interpretable features, or something else entirely?

In this work, we show that these dominant directions---identified by a \emph{spectral edge} in the singular value decomposition of parameter updates---do not correspond to localized structures in parameter or activation space.
Instead, they define \emph{functional modes}: structured perturbations of the model's input-output behavior.
Concretely, each leading singular direction induces a function over the input domain, capturing how the model's internal representations change across inputs.
These functions reveal low-dimensional structure that is invisible to standard interpretability methods.

We demonstrate that conventional approaches fail to capture this structure.
Parameter-space attribution is diffuse across heads and layers, activation perturbations are high-rank, and sparse autoencoder features show no meaningful co-specialization.
This failure is not incidental: it reflects a mismatch between the object of interest and the analysis tools.
The spectral edge identifies directions in parameter space that are coherent at the level of \emph{functions}, not representations.

A key motivation comes from Olah et~al.'s analysis of \emph{interference weights} \citep{olah2025interference}: weights that exist in a network not because they serve any single computation, but as a compromise that allows multiple features to coexist.
Their distinction between ``virtual weights'' (raw magnitudes) and ``effective weights'' (corrected for coactivation) raises a natural question for the spectral edge: are these leading directions genuine functional modes, or interference artifacts?
Our co-usage analysis (\Cref{sec:cousage}) shows that above-edge directions behave as effective weights---coherent functional modes---while below-edge directions carry interference.

When analyzed in input space, these functional modes exhibit striking regularities.
For modular addition, all dominant directions collapse to a single Fourier frequency, revealing a one-dimensional harmonic structure.
For modular multiplication, the same collapse appears only after transforming to the discrete logarithm basis, aligning the analysis with the multiplicative group structure.
In contrast, subtraction spans a small multi-frequency subspace, while $x^2+y^2$ exhibits no single dominant harmonic, instead occupying a structured but non-Fourier subspace.

Crucially, these functional modes are not task-specific artifacts.
In a multitask setting, we find that models reuse functional structure across tasks: the dominant modes of a composite task align with those of simpler constituent tasks.
This provides direct evidence that neural networks learn reusable functional primitives, and that compositional structure emerges through shared training dynamics.

Together, these results suggest a shift in perspective.
Rather than viewing learning in terms of localized circuits or features, we propose that training discovers low-dimensional functional subspaces over the input domain.
Spectral edge dynamics provide a direct probe of these subspaces, revealing the algebraic and compositional structure underlying learned computations.

\paragraph{Main contributions.}

\begin{enumerate}
\item \textbf{Robust spectral edge detection.}
We confirm that a spectral edge---a small set of leading update directions---emerges consistently during grokking and distinguishes grokking from non-grokking conditions across multiple tasks, seeds, and spectral decompositions.

\item \textbf{Negative results for representation-level interpretability.}
We show that standard mechanistic interpretability tools (head attribution, activation-space analysis, sparse autoencoders) fail to capture the spectral edge, indicating that its structure is not localized in parameter or feature space.

\item \textbf{Functional structure in symmetry-adapted bases.}
When analyzed as perturbations over the input domain, spectral edge directions exhibit structured behavior.
For addition and multiplication, this structure collapses to a single Fourier mode in the appropriate group-theoretic basis; for subtraction, it spans a small multi-mode family.

\item \textbf{Non-harmonic functional structure in composed tasks.}
For $x^2+y^2$, no simple one-dimensional Fourier basis captures the spectral edge.
Instead, the structure is partially explained by cross-terms of additive and multiplicative features, indicating a low-dimensional but nontrivially structured functional subspace.

\item \textbf{Evidence of compositional reuse under multitask training.}
In shared-trunk models, the spectral edge of the $x^2+y^2$ head becomes more aligned with additive functional modes characteristic of the addition task, providing direct evidence that multitask training promotes reuse of functional structure.
\end{enumerate}

\section{Experimental Setup}
\label{sec:setup}

\subsection{Model and Tasks}

We use a 2-layer Transformer with $d_\mathrm{model} = 128$, 4 attention heads, $d_\mathrm{ff} = 256$, pre-norm, GELU activation, and $\sim$290k parameters.
Single-task models are trained with AdamW ($\mathrm{lr} = 10^{-3}$, $\mathrm{wd} = 1.0$, $\beta_2 = 0.98$) on modular arithmetic tasks mod $p = 97$ with a 50/50 train-test split.
We study six binary operations: $(a+b)$, $(a-b)$, $(a \cdot b)$, $(a^2+b^2)$, $(a^2+ab+b^2)$, and $(a^3+ab)$ mod~$p$.
The first four grok (achieve $>$95\% test accuracy within 5000 steps); the last two do not, and serve as non-grokking controls.
Each configuration is run with three random seeds ($s \in \{42, 137, 2024\}$).

The functional analysis in Sections~\ref{sec:fourier_results}--\ref{sec:multitask_composition} focuses on the four grokking operations, since the spectral edge is well-defined only when grokking occurs.
The non-grokking operations are used primarily in Section~\ref{sec:gram_gap} as controls.

\subsection{Multitask Models}

We also train shared-trunk models with task-specific classification heads: a dual-task model (add + mul) and a tri-task model (add + mul + $x^2+y^2$), using the same trunk architecture.
The tri-task model groks on all three operations (at steps 17{,}900, 24{,}800, and 30{,}500 respectively), with the staggered ordering providing a natural testbed for compositional reuse \citep{xu2026multitask}.

\subsection{Spectral Edge Definition}
\label{sec:spectral_edge_def}

At each training step $t$, we compute the weight update $\delta\theta_t = \theta_t - \theta_{t-1}$ restricted to attention parameters (4 matrices $\times$ 2 layers, flattened dimension $D = 131{,}072$).
Over a sliding window of $W = 20$ consecutive updates, we form the \emph{Gram matrix}
\[
    G_{ij} = \langle \delta\theta_i,\, \delta\theta_j \rangle, \qquad i,j \in \{t-W+1,\ldots,t\},
\]
and compute its eigenvalues $\sigma_1 \geq \sigma_2 \geq \cdots \geq \sigma_W$.
The \emph{spectral gaps} $g_{k,k+1} = \sigma_k - \sigma_{k+1}$ measure how sharply each eigenvalue separates from the next.
In the BBP (Baik--Ben~Arous--P\'ech\'e) framework \citep{xu2026spectral_edge}, a declining $g_{23} = \sigma_2 - \sigma_3$ indicates that the top-2 block is consolidating: a small number of directions absorb an increasing fraction of update variance while the remaining spectrum compresses.

The \emph{spectral edge position} $k^*$ is defined as the index maximizing the gap ratio $\sigma_k / \sigma_{k+1}$, weighted by cumulative signal mass.
The right singular vectors $v_1, \ldots, v_{k^*}$ of the update matrix constitute the spectral edge directions.

\subsection{Perturbation Response}
\label{sec:perturbation_def}

Given a spectral edge direction $v_k$, we define the \emph{perturbation response} as the change in the model's residual stream under a small displacement along~$v_k$:
\[
    \Delta h_k(a,b) \;=\; h(a,b;\,\theta + \varepsilon v_k) \;-\; h(a,b;\,\theta),
\]
where $h(a,b;\theta)$ denotes the residual stream at position~0 after all encoder layers, and $\varepsilon = 0.005 \cdot \|\theta_\mathrm{attn}\|$.
The scalar field
\[
    f_k(a,b) \;=\; \norm{\Delta h_k(a,b)}^2
\]
captures how sensitive each input pair $(a,b)$ is to perturbation along~$v_k$.
This is the central object of our analysis.
We define a \emph{functional mode} as a spectral edge direction $v_k$ whose induced perturbation field $f_k(a,b)$ exhibits structured variation over the input domain---variation that is organized by the algebraic properties of the task rather than by the architecture of the network.
The direction $v_k$ lives in parameter space, but the functional mode is characterized by $f_k$: a function on the input domain.
The bridge from geometry (singular vectors of weight updates) to function (structured perturbation patterns over inputs) is the main conceptual move of this paper.

\subsection{Fourier Analysis of the Perturbation Response}
\label{sec:fourier_method}

For a given grouping variable $q(a,b) \in \{0, \ldots, M-1\}$, we compute the group-averaged perturbation signal
\[
    \bar{f}_k[q] \;=\; \frac{1}{|\{(a,b) : q(a,b) = q\}|} \sum_{q(a,b)=q} f_k(a,b),
\]
an $M$-point signal indexed by the group elements.
We then compute its discrete Fourier transform.
The \emph{Fourier concentration} $F_k$ is the fraction of total power at the peak frequency:
\[
    F_k \;=\; \max_\omega \frac{|\hat{\bar{f}}_k[\omega]|^2}{\sum_{\omega'} |\hat{\bar{f}}_k[\omega']|^2}.
\]
The uniform baseline is $F = 2/M$ (one cosine--sine pair out of $M/2$).
A high $F_k$ indicates that the perturbation pattern is dominated by a single Fourier mode of the grouping variable.

The choice of grouping variable---and hence the Fourier basis---is the key degree of freedom.
Different choices correspond to different group-theoretic decompositions of the input domain.

\section{Functional Subspaces: Definitions and Framework}
\label{sec:theory}

Before presenting empirical results, we formalize the key concepts.
This section separates what is definitional (the mapping from parameter directions to functions), what is empirically observed (the low-dimensional structure), and what is conjectured (the relationship to group representations).

\subsection{Definitions}

Let $\theta \in \R^D$ denote model parameters and $h(x;\theta) \in \R^{d_\mathrm{model}}$ the residual stream at position~0 for input $x = (a,b)$.
Let $v_k \in \R^D$ be a spectral edge direction.

\begin{quote}
\textbf{Definition 1} (Functional mode).
The \emph{induced activation perturbation} along $v_k$ is
\[
    \Delta h_k(x) \;:=\; \left.\frac{d}{d\varepsilon}\, h(x;\, \theta + \varepsilon v_k)\right|_{\varepsilon=0},
\]
and the corresponding \emph{scalar functional mode} is
\[
    f_k(x) \;:=\; \norm{\Delta h_k(x)}^2.
\]
Although $v_k$ is a direction in parameter space $\R^D$, the induced function $f_k$ lives in input space $\mathcal{X} = \{0,\ldots,p-1\}^2$.
In practice, we approximate the derivative with a finite difference at $\varepsilon = 0.005 \cdot \norm{\theta_\mathrm{attn}}$ (Section~\ref{sec:perturbation_def}).
\end{quote}

\begin{quote}
\textbf{Definition 2} (Functional subspace).
Let $v_1, \ldots, v_{k^*}$ be the spectral edge directions.
The \emph{functional subspace} induced by spectral edge dynamics is
\[
    \mathcal{F} \;:=\; \mathrm{span}\{f_1, \ldots, f_{k^*}\} \;\subset\; L^2(\mathcal{X}).
\]
\end{quote}

\noindent
These definitions are exact: they introduce a mapping from parameter-space geometry to function space.
The content of the paper is what this mapping reveals empirically.

\subsection{Empirical Observation}

We observe a consistent separation of structure across spaces:

\begin{quote}
\textbf{Empirical Observation} (Functional structure of the spectral edge).
During grokking, weight updates concentrate onto a parameter subspace whose induced functional subspace $\mathcal{F}$ is low-dimensional and structured.
Specifically:
\begin{itemize}
    \item $\Delta h_k(x)$ is high-rank in activation space (effective rank $\approx 40$ out of $d_\mathrm{model} = 128$);
    \item but $f_k(x)$ exhibits low-dimensional structure over inputs, concentrated on a small number of Fourier modes when expressed in the correct group-theoretic basis.
\end{itemize}
The structure of learning is not in how the perturbation is distributed across neurons, but in \emph{which inputs} are most affected.
\end{quote}

\noindent
This observation is documented across four grokking operations, three seeds, and three independent spectral decompositions (Sections~\ref{sec:gram_gap}--\ref{sec:fourier_results}).

\subsection{Conjecture}

The empirical observations suggest two conjectures of different strength.

\begin{quote}
\textbf{Conjecture 1} (Dynamical selection of task-relevant eigenmodes).
The spectral edge of the weight-update trajectory preferentially selects directions whose functional modes $f_k$ align with the natural eigenmodes of the task.
For tasks that factor through a group homomorphism $\phi : \mathcal{X} \to G$, these eigenmodes are the irreducible representations (characters) of $G$.
Among all possible update directions, those with the largest singular values induce perturbation patterns concentrated on the same characters that the model uses to solve the task.
\end{quote}

\noindent
The non-trivial content here is not that the model uses group characters---that is expected from prior work \citep{nanda2023grokking} and follows from the task structure.
The conjecture is about the \emph{training dynamics}: that SGD concentrates its updates along directions aligned with the task-relevant eigenmodes, rather than distributing updates uniformly across all possible functional directions.
This is a statement about loss landscape geometry, not representation theory.

\paragraph{Why Fourier, specifically?}
The Fourier basis is not claimed to be universal.
It is the specific realization of the task's eigenmodes for finite abelian groups: characters diagonalize convolution on these groups, and are the natural basis in which group-theoretic computations decompose.
For tasks with different structure---non-abelian groups, continuous symmetries, or no obvious algebraic structure---the relevant eigenmodes would be different (irreducible representations, spherical harmonics, or eigenfunctions of some task-dependent operator).
What may be general is not ``Fourier'' but the principle that training dynamics align with whatever eigenmodes govern the task.
Weight decay likely plays a role in this selection: Fourier representations are parameter-efficient for group-theoretic tasks (they exploit symmetry), so regularization biases the optimizer toward these solutions.
This is consistent with the observation that the spectral edge forms only under weight decay ($\mathrm{wd} = 1.0$), not without it.

\begin{quote}
\textbf{Conjecture 2} (Compositional inheritance under shared training).
Let $T$ be a task whose computation can be decomposed into simpler operations $T_2, T_3$ (e.g., $a^2+b^2 = (a+b)^2 - 2ab$ decomposes into addition and multiplication).
Denote by $\mathcal{F}_T^{\mathrm{shared}}$ the functional subspace of $T$ when trained jointly with $T_2$ and $T_3$ in a shared-trunk model, and by $\mathcal{F}_T^{\mathrm{single}}$ the subspace when trained alone.
Then shared training promotes inheritance of component modes:
\[
    \mathcal{F}_T^{\mathrm{shared}} \;\cap\; (\mathcal{F}_{T_2} + \mathcal{F}_{T_3}) \;\supsetneq\; \mathcal{F}_T^{\mathrm{single}} \;\cap\; (\mathcal{F}_{T_2} + \mathcal{F}_{T_3}).
\]
That is, shared training increases the overlap between the composite task's functional modes and those of its components.
Under single-task training, the model is free to learn an opaque representation with little or no overlap; under shared-trunk training, functional modes are reused.

Note that this conjecture addresses only whether the component modes $\mathcal{F}_{T_2}, \mathcal{F}_{T_3}$ are inherited---it does not characterize the full functional subspace $\mathcal{F}_T$, which may also contain modes arising from how the components are combined (e.g., the squaring step in $(a+b)^2 - 2ab$).
Our cross-term probing captures some of this interaction but does not fully explain it ($R^2 = 0.16$).
\end{quote}

\noindent
We do not prove either conjecture.
The remainder of the paper provides evidence for Conjecture~1 across four tasks (two clean cases, one partial, one negative) and for Conjecture~2 via single-task vs.\ multitask comparison on $x^2+y^2$.
We are explicit about where the evidence is strong, where it is weak, and where it breaks down.

\paragraph{What the conjectures do not predict.}
Neither conjecture predicts \emph{which} irreps are selected or \emph{how many}.
For modular addition and multiplication, we observe $m = 1$ (a single dominant character); for subtraction, $m \approx 3$; these tasks all factor through the same group $\Z/97\Z$.
A stronger theory would predict $m$ from properties of the target function (e.g., the number of Fourier components needed for $\varepsilon$-approximation of the loss landscape).
We leave this as an open problem.

\section{The Spectral Edge Exists and Discriminates Grokking}
\label{sec:gram_gap}

\subsection{Gram Matrix Gap Dynamics}

We track the spectral gap $g_{23} = \sigma_2 - \sigma_3$ of the Gram matrix over training.
A declining $g_{23}$ indicates that the top-2 eigenvalues are separating from the rest: the update trajectory is concentrating onto a lower-dimensional subspace.

\begin{table}[t]
\centering
\caption{Spectral edge discrimination: $g_{23}$ decline magnitude (ratio of early-training to post-grok values) across 36 single-task conditions.
Grokking runs show $15$--$110\times$ decline; non-grokking conditions show $<2\times$.}
\label{tab:gap_discrimination}
\begin{tabular}{lcccc}
\toprule
\textbf{Condition} & \textbf{Runs} & \textbf{$g_{23}$ decline detected} & \textbf{Mean magnitude} \\
\midrule
Grokking ($\mathrm{wd}=1.0$, 4 ops) & 12 & 12/12 (100\%) & $40\times$ \\
Control ($\mathrm{wd}=0$, 4 ops) & 12 & 0/12 (0\%) & $1.2\times$ \\
Non-grokking ops ($\mathrm{wd}=1.0$) & 6 & 1/6 (17\%) & $1.5\times$ \\
Non-grokking ops ($\mathrm{wd}=0$) & 6 & 0/6 (0\%) & $1.1\times$ \\
\bottomrule
\end{tabular}
\end{table}

Across 36 single-task conditions (6~operations $\times$ 2~weight-decay settings $\times$ 3~seeds), the $g_{23}$ gap cleanly separates grokking from non-grokking regimes (\Cref{tab:gap_discrimination}).
All 12 grokking runs show $g_{23}$ decline with magnitude $15$--$110\times$.
Only 1 of 24 control conditions shows comparable decline (a late anomalous event at step 41{,}700 in a non-grokking operation).


\subsection{Three Independent Decompositions Agree}

To verify that the spectral edge is not an artifact of the Gram matrix construction, we compare three independent spectral analyses on the same training trajectories:
\begin{enumerate}
    \item \textbf{Gram matrix} (sliding-window update SVD): eigenvalues of $G_{ij}$ as defined above.
    \item \textbf{Displacement PCA} (expanding-window from initialization): eigenvalues of the covariance of $\theta_t - \theta_0$.
    \item \textbf{Weight-matrix SVD}: singular values $\sigma_1 \geq \sigma_2 \geq \cdots$ of the individual attention matrices $W_Q$, $W_K$.
\end{enumerate}

All three yield consistent temporal signatures: the spectral edge forms during the memorization phase, sharpens at the grokking transition, and stabilizes post-grok.
The weight-matrix SVD additionally reveals a mechanism: grokking is preceded by a transient near-degeneracy of the leading singular values ($\sigma_1 \approx \sigma_2$), followed by symmetry breaking ($\sigma_1 \gg \sigma_2$) at generalization.
This near-degeneracy creates an orientation instability in the top singular subspace, consistent with the sharp rise in commutator defects observed in prior work \citep{xu2026lowdim}.

\section{The Spectral Edge Lives in Function Space, Not Representation Space}
\label{sec:mech_fail}

Standard mechanistic interpretability operates in \emph{representation space}: it decomposes networks into structural units (heads, neurons) or learned features (SAE dictionary elements).
We test whether the spectral edge can be understood in these terms.
Each test targets a natural hypothesis; each fails---not because the edge lacks structure, but because these tools operate in the wrong space.

\paragraph{Hypothesis 1: Head localization.}
If $v_k$ corresponded to a single attention head, its parameter mass would concentrate in one (layer, head, matrix-type) block.
We compute the \emph{head purity}: $\max_{\ell,h,m} \|v_k[\text{block}_{\ell,h,m}]\|^2 / \|v_k\|^2$, where the maximum runs over 2~layers $\times$ 4~heads $\times$ 4~matrix types.
Across all four grokking operations, head purity $\approx 0.14$ for all top-3 directions---barely above the uniform baseline $1/8 = 0.125$ for the 8 layer$\times$head combinations (marginalizing over matrix type).
\textbf{The spectral edge is globally distributed across all heads.}

\paragraph{Hypothesis 2: Low-rank activation feature.}
Even if $v_k$ is diffuse in parameter space, it might produce a low-rank perturbation in activation space.
We compute the \emph{effective rank} of the displacement matrix $[\Delta h_k(x_1), \ldots, \Delta h_k(x_N)]^\top \in \R^{N \times d_\mathrm{model}}$, defined as $\exp(H(p))$ where $H(p) = -\sum_j p_j \log p_j$ is the entropy of the normalized squared singular values.
Result: effective rank $\approx 40$ for all $v_k$ and all operations, out of $d_\mathrm{model} = 128$.
\textbf{The perturbation is diffuse in activation space.}

However, the \emph{Fourier peakedness} of the scalar field $f_k(a,b) = \|\Delta h_k\|^2$ reaches $32.8\times$ uniform for $v_1$ at the grokking transition (addition, $\omega = 25$).
This indicates that while $\Delta h_k$ is high-dimensional as a vector in $\R^{d_\mathrm{model}}$, the \emph{pattern} of where the perturbation is large across inputs has sharp structure.

\paragraph{Hypothesis 3: Shared sparse features.}
We train a TopK sparse autoencoder ($d_\mathrm{SAE} = 512$, $k = 32$) on the post-grok residual stream and compute the Jaccard overlap of the top-20 most-affected SAE features between pairs of top-3 directions.
Observed Jaccard ranges from 0.25 (addition) to 0.60 (subtraction).
We test against three null models:
\begin{itemize}
    \item \textbf{Combinatorial null}: expected Jaccard $\approx 0.02$ for random draws of 20 from 512.
    \item \textbf{Random-direction null} (50 random unit vectors in parameter space): mean Jaccard ranges from 0.30 to 0.83, depending on operation.
    \item \textbf{Angle-matched null} (synthetic direction pairs with the same $45^\circ$ subspace angle as the observed top-3): $p \geq 0.97$ for all operations.
\end{itemize}
The observed Jaccard values fall \emph{at or below} the random-direction mean in all cases.
\textbf{The apparent SAE overlap is an artifact of representation density at this sparsity level, not a property of the spectral edge.}

\paragraph{Diagnosis: category mismatch, not absence of structure.}
These three null results do not mean the spectral edge is unstructured.
They mean that its structure is invisible to tools that operate in representation space---heads, activation features, and sparse dictionaries.
The Fourier peakedness result ($32.8\times$ uniform) already reveals that the structure exists, but in a different space: the pattern of how $f_k(a,b)$ varies \emph{across inputs}.
The spectral edge is a functional object.
The next section shows what it looks like when analyzed in the correct functional coordinates.

\section{Functional Structure in Symmetry-Adapted Bases}
\label{sec:fourier_results}

\subsection{From Single Modes to Functional Subspaces}

We initially hypothesized that the leading update directions $\{v_1, v_2, v_3\}$ would align with a single dominant Fourier mode.
While this holds for some tasks, the full picture is more nuanced.
Across all operations, the spectral edge consistently identifies a \emph{low-dimensional functional structure}.
In some cases this collapses to a single dominant mode; in others it spans a small family of modes.

This leads to a refined view: spectral edge dynamics select a \emph{functional subspace}, not necessarily a single frequency.

\subsection{Addition: Collapse to a Single Mode}
\label{sec:fourier_add}

For modular addition, we set the grouping variable to $q = (a+b) \bmod p$, so the Fourier basis consists of additive characters $\chi_\omega(a,b) = e^{2\pi i\omega(a+b)/p}$ with $\omega \in \{1, \ldots, 48\}$.

\begin{table}[t]
\centering
\caption{Fourier profiles of spectral edge directions for modular addition.
All three top directions concentrate on $\omega \approx 25$--$26$.
Bulk directions (below the spectral edge) show lower and less consistent concentration.}
\label{tab:add_fourier}
\begin{tabular}{lccc}
\toprule
\textbf{Direction} & \textbf{Peak $\omega$} & \textbf{Concentration $F$} & \textbf{Top-3 $\omega$} \\
\midrule
$v_1$ (edge) & 26 & 0.33 & $\{26, 12, 25\}$ \\
$v_2$ (edge) & 26 & 0.16 & $\{26, 8, 14\}$ \\
$v_3$ (edge) & 25 & 0.40 & $\{25, 26, 3\}$ \\
\midrule
$v_4$ (bulk) & 8 & 0.18 & $\{8, 26, 11\}$ \\
$v_5$ (bulk) & 26 & 0.12 & $\{26, 8, 15\}$ \\
\midrule
\textit{uniform baseline} & --- & 0.021 & --- \\
\bottomrule
\end{tabular}
\end{table}

The result is maximally simple (\Cref{tab:add_fourier}):
\begin{itemize}
    \item All three leading directions concentrate on the same frequency $\omega \approx 25$--$26$,
    \item Fourier concentration reaches $F = 0.40$ (baseline $0.021$, a $19\times$ elevation),
    \item The top-3 mean $F = 0.29$ exceeds the bulk mean $F = 0.18$ by $1.6\times$.
\end{itemize}
Per-PC analysis of the displacement vector $\Delta h_1$ reveals even sharper concentration: the first principal component reaches $F = 0.67$ at $\omega = 26$.
2D Fourier analysis (computing $P(\omega_a, \omega_b)$ over individual inputs $a$ and $b$) shows power concentrated on the \emph{diagonal} $(\omega_a, \omega_b) = (26, 26)$, consistent with the additive structure $e^{2\pi i\omega(a+b)/p} = e^{2\pi i\omega a/p} \cdot e^{2\pi i\omega b/p}$.

The top block effectively collapses to a one-dimensional functional subspace: $\mathrm{span}\{g_1, g_2, g_3\} \approx \mathrm{span}\{\chi_{25\text{--}26}\}$.

\paragraph{A note on signal strength.}
We do not claim exact decomposition.
$F = 0.40$ means 60\% of the power is distributed across other frequencies; the perturbation response is a noisy, distributed object.
The key signal is not purity but \emph{basis alignment}: consistent frequency selection across all top-3 directions, and the $19\times$ elevation over the uniform baseline.
The strongest evidence comes from relative comparisons---the $5.9\times$ improvement for multiplication under discrete-log (\Cref{sec:fourier_mul}), and the fact that applying the wrong basis actively destroys structure.

\begin{figure}[t]
    \centering
    \includegraphics[width=\textwidth]{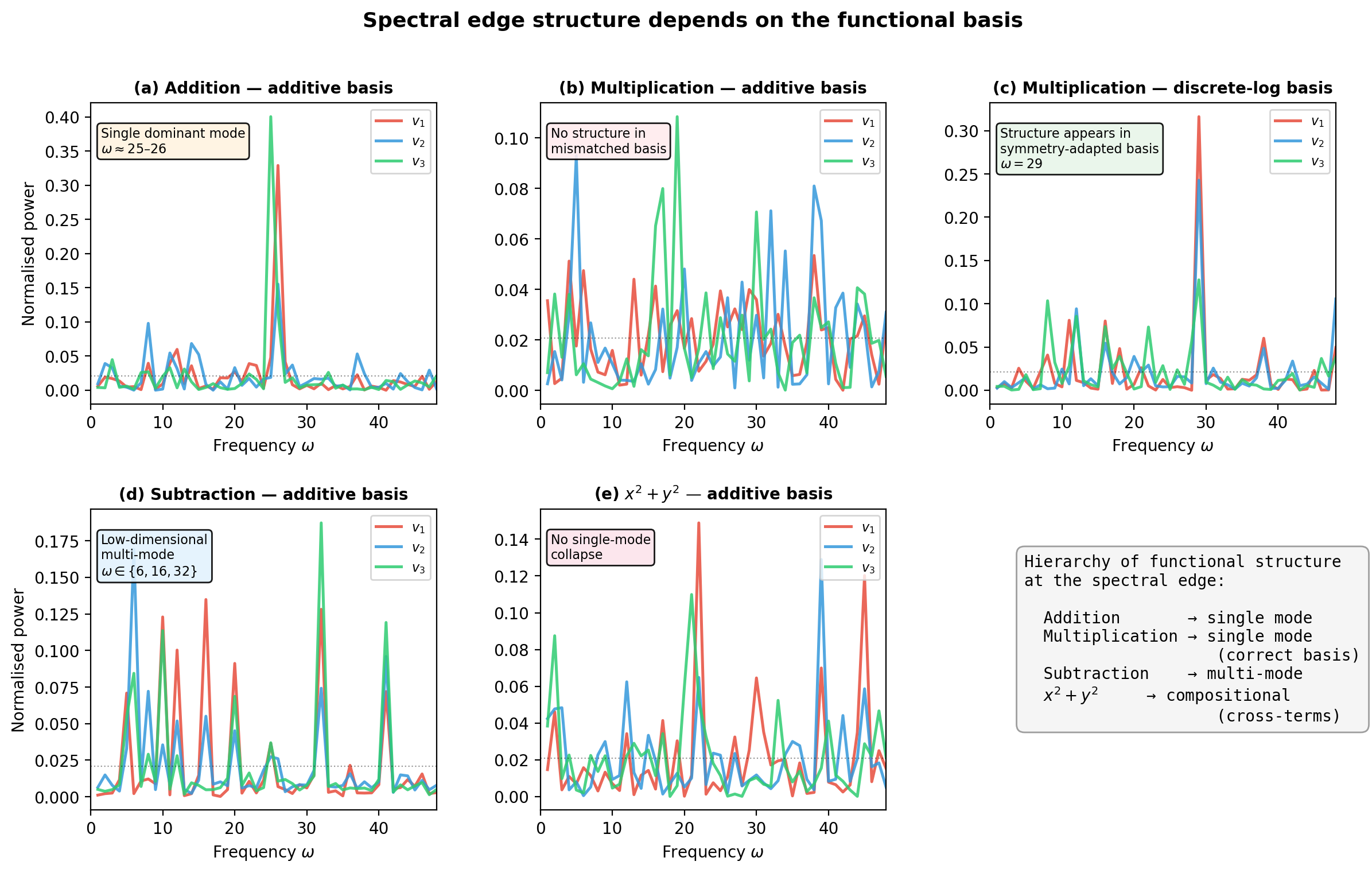}
    \caption{The spectral edge consistently identifies low-dimensional functional structure, but its simplicity depends on the basis.
    \textbf{(a)}~Addition collapses to a single Fourier mode ($\omega \approx 25$--$26$) under the additive basis.
    \textbf{(b)}~Multiplication shows no structure in the additive basis.
    \textbf{(c)}~The same multiplication data collapses to $\omega = 29$ under the discrete-log basis ($5.9\times$ concentration improvement for $v_1$).
    \textbf{(d)}~Subtraction spans a small set of modes ($\omega \in \{6, 16, 32\}$).
    \textbf{(e)}~$x^2+y^2$ exhibits distributed structure not captured by any single one-dimensional harmonic basis.
    Dashed line: uniform baseline.
    All spectra computed on post-grok checkpoints, seed 42.}
    \label{fig:basis_comparison}
\end{figure}

\subsection{Multiplication: Collapse in a Symmetry-Adapted Basis}
\label{sec:fourier_mul}

For multiplication, the same collapse appears---but only after transforming to the appropriate group-theoretic basis.

In the additive Fourier basis (grouping by $(a \cdot b) \bmod p$), the signal is weak and diffuse: peak $F \leq 0.11$ for all top-3 directions, with scattered frequencies ($\omega \in \{5, 19, 38\}$).
The multiplicative group $(\Z/p\Z)^*$ has a different character theory.
Given a primitive root $g$ (here $g = 5$ for $p = 97$), the natural grouping variable is $q = (\mathrm{dlog}_g(a) + \mathrm{dlog}_g(b)) \bmod (p-1)$, with period $p-1 = 96$.

Under this discrete-log basis:
\begin{itemize}
    \item All three leading directions collapse onto $\omega = 29$,
    \item Concentration increases by up to $5.9\times$ ($v_1$: $F = 0.054 \to 0.32$),
    \item Mean top-3 improvement: $2.7\times$.
\end{itemize}
As a control, addition's Fourier concentration \emph{decreases} under the discrete-log basis ($0.2$--$0.3\times$ of its additive-basis value), confirming that the improvement is basis-specific.

This demonstrates that the collapse to a single mode depends on using a \emph{symmetry-adapted basis} matched to the algebraic structure of the task.
The simplicity of the spectral edge is not intrinsic to parameter space; it depends on expressing the perturbation response in the correct functional coordinates.

\subsection{Subtraction: A Small Family of Modes}
\label{sec:fourier_sub}

For subtraction, the additive Fourier basis (grouping by $(a-b) \bmod p$) remains group-theoretically appropriate, but the structure does not collapse to a single frequency.
Instead:
\begin{itemize}
    \item Each leading direction aligns with a distinct frequency: $v_1 \to \omega = 16$, $v_2 \to \omega = 6$, $v_3 \to \omega = 32$,
    \item The union of their top-3 frequency sets spans five modes $\{6, 10, 16, 32, 41\}$.
\end{itemize}
The spectral edge spans a small multi-mode subspace:
$\mathrm{span}\{g_1, g_2, g_3\} \subset \mathrm{span}\{\chi_6, \chi_{16}, \chi_{32}\}$.

However, the top-3 concentration ($F \approx 0.16$) does not separate from the bulk ($F \approx 0.16$): unlike addition and multiplication, the spectral edge directions are not more Fourier-concentrated than bulk directions below the edge.
This suggests either that the functional subspace is not well-aligned with a single harmonic basis, or that our analysis fails to identify the appropriate basis for subtraction.
We report this as an honest negative: the multi-mode description is a plausible interpretation but is not supported by a clean separation from the null.

\subsection{$x^2 + y^2$: Low-Dimensional Structure Is Not the Same as Harmonic Structure}
\label{sec:fourier_x2y2}

The quadratic task provides the deepest conceptual test.
Addition and multiplication show that the spectral edge can collapse to a single Fourier mode; one might conclude that spectral edge $=$ Fourier mode.
The $x^2+y^2$ case refutes this identification: the spectral edge remains low-dimensional and meaningful, but \emph{no} simple harmonic basis captures it.
This is the case that forces us beyond Fourier modes to the broader concept of low-dimensional functional subspaces.

We test five distinct Fourier bases:
\begin{enumerate}[label=(\alph*)]
    \item \textbf{Additive}: group by $(a^2+b^2) \bmod p$, period $p = 97$.
    \item \textbf{Multiplicative}: $\mathrm{dlog}$ of the output, period $p-1 = 96$.
    \item \textbf{Gaussian integer}: exploiting $a^2+b^2 \equiv uv \pmod{p}$ where $u = a + i_p b$, $v = a - i_p b$, and $i_p = \sqrt{-1} \bmod p$ (here $i_p = 22$, since $22^2 \equiv -1 \pmod{97}$).
    Grouping by $(\mathrm{dlog}(u) + \mathrm{dlog}(v)) \bmod (p-1)$.
    \item \textbf{Angular}: $(\mathrm{dlog}(u) - \mathrm{dlog}(v)) \bmod (p-1)$, corresponding to the phase of the Gaussian integer $z = a + i_p b$.
    \item \textbf{2D Fourier}: $P(\omega_a, \omega_b)$ in both standard and squared-input coordinates.
\end{enumerate}
None yields a dominant mode.
Peak concentration remains $F \approx 0.10$--$0.15$ across all bases, comparable to bulk directions below the spectral edge.

The 2D Gaussian Fourier analysis is informative in a different way: it reveals 97\% off-diagonal power in the $(\mathrm{dlog}(u), \mathrm{dlog}(v))$ plane, indicating that the perturbation is sensitive to the \emph{angular} coordinate (which pair $(a,b)$ maps to a given output) rather than the \emph{radial} coordinate (the output value itself).
But even the angular variable does not concentrate on a single frequency.

\paragraph{Composition test.}
The algebraic identity $a^2+b^2 = (a+b)^2 - 2ab$ suggests that the model may \emph{compose} addition and multiplication.
We test this by probing the perturbation with three feature sets via ridge regression:
(A)~additive Fourier features ($\cos(2\pi\omega(a+b)/p)$ for all $\omega$),
(B)~multiplicative (discrete-log) features, and
(C)~their cross-terms (products of additive and multiplicative features).

\begin{table}[t]
\centering
\caption{Multivariate probe $R^2$ (predicting the full displacement $\Delta h_k$ from Fourier features).
For addition and multiplication, the corresponding single-basis features explain substantial variance.
For $x^2+y^2$, neither alone suffices, but cross-terms provide a $4\times$ boost.}
\label{tab:composition}
\begin{tabular}{lccccc}
\toprule
\textbf{Task} & \textbf{Add features} & \textbf{Mul features} & \textbf{Combined} & \textbf{+Cross-terms} \\
\midrule
Addition & 0.17--0.29 & 0.02 & 0.18--0.31 & 0.29--0.40 \\
Multiplication & 0.02 & 0.11--0.26 & 0.12--0.27 & 0.23--0.37 \\
$x^2+y^2$ & 0.01 & 0.02 & 0.03--0.04 & \textbf{0.15--0.16} \\
\bottomrule
\end{tabular}
\end{table}

The results (\Cref{tab:composition}) are consistent with the composition hypothesis:
addition is explained by additive features alone; multiplication by dlog features alone.
For $x^2+y^2$, neither suffices ($R^2 \leq 0.02$), but cross-terms raise $R^2$ from 0.04 to 0.16---a $4\times$ increase.

This $R^2 = 0.16$ accounts for only a fraction of the total variance, indicating that the composition is not the whole story.
But the qualitative pattern---that cross-terms help specifically for $x^2+y^2$ and not for the simpler operations---supports the compositional interpretation.

\subsection{Functional Modes Are Reusable: Multitask Evidence}
\label{sec:multitask_composition}

The composition test above shows that $x^2+y^2$ involves both additive and multiplicative structure.
A stronger test asks: if the model is \emph{forced} to share parameters with addition and multiplication (via a shared trunk), does the $x^2+y^2$ spectral edge inherit their functional modes?
If so, this would be evidence that functional modes are not just descriptive labels but \emph{reusable building blocks} of learned computation.

\begin{table}[t]
\centering
\caption{Single-task vs.\ tritask comparison for $x^2+y^2$.
The tritask model shows stronger alignment with additive modes and larger composition synergy.}
\label{tab:multitask}
\begin{tabular}{lccc}
\toprule
\textbf{Metric} & \textbf{Single-task} & \textbf{Tritask} & \textbf{Ratio} \\
\midrule
Additive concentration at $\omega=26$ & $F = 0.09$ & $F = 0.20$ & $2.3\times$ \\
Composition synergy ($v_1$) & $+0.024$ & $+0.043$ & $1.7\times$ \\
Peak additive frequency & $\omega = 17$ (scattered) & $\omega = 26$ (= addition mode) & --- \\
\bottomrule
\end{tabular}
\end{table}

The results (\Cref{tab:multitask}) support the composition hypothesis:
\begin{itemize}
    \item Additive Fourier concentration at $\omega = 26$ increases $2.3\times$ (from $F = 0.09$ to $F = 0.20$).
    \item Composition synergy (combined $R^2$ minus $\max(\text{add}, \text{mul})$ individually) increases $1.7\times$.
    \item Crucially, the tritask spectral edge inherits the addition circuit's characteristic frequency $\omega = 26$, which is absent in the single-task model.
\end{itemize}
The shared trunk forces the $x^2+y^2$ computation to partially reuse the addition and multiplication functional modes, and the spectral edge makes this reuse visible.

This result elevates functional modes from a descriptive tool to a potentially causal one: the same modes that organize simple tasks can be recruited by more complex tasks, suggesting that functional modes are reusable primitives of learned computation.

\subsection{Unified Picture}
\label{sec:hierarchy}

\begin{table}[t]
\centering
\caption{Hierarchy of functional structure at the spectral edge.
The correct basis and the nature of the functional subspace depend on the algebraic structure of the task.}
\label{tab:hierarchy}
\begin{tabular}{llll}
\toprule
\textbf{Task} & \textbf{Basis} & \textbf{Structure} & \textbf{Peak $F$} \\
\midrule
$(a+b) \bmod p$ & Additive characters & Single mode ($\omega \approx 25$--$26$) & 0.40 \\
$(a \cdot b) \bmod p$ & Discrete-log characters & Single mode ($\omega = 29$) & 0.32 \\
$(a-b) \bmod p$ & Additive characters & Multi-mode ($\omega \in \{6, 16, 32\}$) & 0.19 \\
$(a^2\!+\!b^2) \bmod p$ & Cross (add $\times$ mul) & Compositional & $0.16^\dagger$ \\
\bottomrule
\end{tabular}

\smallskip
\noindent
{\footnotesize $^\dagger$No single-basis $F$; value is multivariate probe $R^2$ from \Cref{tab:composition}.}
\end{table}

These results (\Cref{tab:hierarchy}) suggest the following interpretation:

\begin{quote}
\emph{Spectral edge dynamics identify a low-dimensional functional subspace.
In some tasks, this subspace collapses to a single harmonic mode; in others, it spans a small set of modes or requires richer bases to describe.}
\end{quote}

\noindent
Fourier simplicity is not universal, but low-dimensional functional structure appears to be.

\section{Connection to Interference Weights}
\label{sec:cousage}

The interference weight framework of \citet{olah2025interference} provides a complementary lens on the spectral edge.
For each test input $x$ and direction $v_k$, we compute the per-example activation $a_k(x) = \langle \nabla_\theta \ell(x),\, v_k \rangle$, measuring how much the gradient for input~$x$ aligns with~$v_k$.

Post-grok, above-edge directions ($k \leq k^*$) are activated by coherent, non-overlapping subsets of inputs, while below-edge directions are activated diffusely with high inter-direction correlation.
In Olah et~al.'s terminology, above-edge directions behave as \emph{effective weights} (virtual weight $\approx$ effective weight after coactivation correction), while below-edge directions carry \emph{interference}---parameter mass that exists as a compromise between competing computations.

In multitask models, above-edge directions also show high cross-task sharedness ($S \approx 0.8$--$0.9$), while below-edge directions are more task-specific ($S \approx 0.3$--$0.5$).
This is consistent with the Fourier picture: shared functional modes (such as $\omega = 26$, used by both addition and the addition subcircuit of $x^2+y^2$) appear as above-edge directions, while task-specific refinements concentrate in the bulk.
Full details of the co-usage analysis are in \Cref{sec:app_cousage}.

\section{Discussion}
\label{sec:discussion}

\subsection{A Different Level of Description}

Our results suggest that the spectral edge occupies a different level of description from standard mechanistic interpretability:

\begin{center}
\begin{tabular}{ll}
\toprule
\textbf{Approach} & \textbf{Objects} \\
\midrule
Mechanistic interpretability & neurons, heads, circuits, features \\
This work & \emph{functional modes over the input domain} \\
\bottomrule
\end{tabular}
\end{center}

\noindent
The category mismatch between representation-level tools and function-level objects is not incidental.
It reflects a genuine difference in what is being described: the spectral edge does not say ``which neuron fires,'' but ``which function over inputs is being learned.''
This is closer to harmonic analysis or operator theory than to circuit-level decomposition.

Prior work \citep{nanda2023grokking} identifies Fourier structure in the final trained representations.
Our results show that the same structure appears in the training dynamics, via spectral edge directions.
Circuit analysis describes the destination; spectral edge analysis describes the trajectory.
That the same frequencies appear in both views suggests that Fourier modes are not merely properties of the final representation, but are \emph{selected during optimization}.

\subsection{What We Claim and What We Do Not}

We claim:
\begin{itemize}
    \item Training dynamics select low-dimensional functional modes over the input domain.
    \item These modes align with Fourier bases when the task admits appropriate group structure.
    \item They are reusable across tasks under shared training.
    \item Standard representation-level tools cannot see them (category mismatch).
\end{itemize}

\noindent
We do \emph{not} claim:
\begin{itemize}
    \item Exact decomposition (our Fourier concentrations are moderate: $F \leq 0.40$; our probe $R^2 \leq 0.16$ for $x^2+y^2$).
    \item Universality beyond modular arithmetic.
    \item A complete theory (the conjecture in \Cref{sec:theory} is not proven).
    \item That subtraction's functional subspace is well-characterized (our analysis does not achieve edge-bulk separation for this task).
\end{itemize}

\subsection{Limitations}

Our analysis is restricted to modular arithmetic tasks in small transformer models, where the underlying algebraic structure is relatively transparent.
Extending this framework to domains such as language, where the appropriate functional basis is not obvious \textit{a priori}, remains an open challenge.
In symmetry-rich tasks, the correct character basis can reveal functional structure; more complex tasks may require \emph{discovering} an appropriate functional basis rather than assuming one.

Two specific extensions are immediate: a systematic multitask ablation (training $x^2+y^2$ with only addition, only multiplication, or both) to isolate which subcircuit donates which functional mode, and an investigation of whether the spectral edge structure persists in deeper models or different task families.

\paragraph{The critical test.}
The value of the functional-mode framework depends on whether it extends beyond settings where the algebraic structure is known \textit{a priori}.
In modular arithmetic, we can identify the correct Fourier basis because we know the symmetry group.
In language, vision, or reasoning tasks, the relevant ``eigenmodes'' are unknown---and discovering them is the hard problem.

The most impactful next step would be to show that the spectral edge of a language model's training trajectory aligns with \emph{some} interpretable functional basis---syntactic structures, semantic roles, or task-relevant features---even without knowing the basis in advance.
If functional modes exist in realistic settings, they could provide a new handle on questions that the field currently approaches through scaling laws and probing classifiers: \emph{what} is being learned at each stage of training, not just \emph{how much}.
If they do not, then this work characterizes an interesting property of group-theoretic tasks that does not generalize.
We consider this an open empirical question.

\section{Related Work}
\label{sec:related}

\paragraph{Grokking and Fourier representations.}
\citet{power2022grokking} first observed delayed generalization in modular arithmetic.
\citet{nanda2023grokking} identified Fourier-basis representations (``grokking circuits'') in 1-layer models trained on modular addition, showing that the trained network uses specific Fourier frequencies (e.g., $\omega = 25$ for mod-97 addition) in its internal computation.
\citet{zhong2024clock} described clock and pizza representations in this setting.
\citet{chughtai2023toy} and \citet{stander2024grokking} studied group-theoretic structure in grokked models.

Our work is complementary to this line of research.
Circuit analysis identifies Fourier structure in the \emph{final trained weights}---describing the destination.
Spectral edge analysis identifies the same Fourier structure in the \emph{training dynamics}---describing the trajectory.
That the same frequencies ($\omega \approx 25$--$26$ for addition, $\omega = 29$ in the dlog basis for multiplication) appear in both views suggests that Fourier modes are not merely properties of the final representation, but are \emph{selected during optimization}.
The spectral edge provides a dynamical lens on what circuit analysis observes statically.

\paragraph{Spectral edge and training dynamics.}
\citet{xu2026lowdim} showed that weight trajectories during grokking lie on low-dimensional execution manifolds, and that commutator defects in the normal bundle predict the generalization transition.
\citet{xu2026multitask} extended this to multitask settings, analyzing transverse instability and superposition in shared-trunk models.
\citet{xu2026spectral_edge} formalized the spectral edge via Gram matrix analysis and BBP theory.
We extend these geometric results by characterizing the \emph{functional content} of the spectral edge directions.

\paragraph{Mechanistic interpretability.}
Sparse autoencoders \citep{bricken2023monosemanticity, cunningham2024sparse} and head/circuit attribution \citep{elhage2021mathematical} provide structural decompositions of neural network internals.
The superposition framework \citep{elhage2022superposition} describes how models represent more features than they have dimensions.
\citet{olah2025interference} extend this by distinguishing virtual weights (raw magnitudes) from effective weights (corrected for coactivation), showing that much of the apparent structure in weight matrices is interference rather than computation.
This distinction directly inspired our co-usage analysis: the spectral edge position $k^*$ marks the boundary where virtual weights transition from effective (above-edge) to interference-dominated (below-edge).
Our negative results for SAE and head attribution highlight a limitation of structural interpretability tools when applied to distributed, parameter-space objects; our Fourier analysis provides the functional basis that these tools lack.

\paragraph{Intrinsic dimensionality.}
\citet{li2018measuring} showed that neural network optimization occurs in low-dimensional subspaces.
Our Fourier analysis characterizes the \emph{functional content} of these subspaces, moving beyond dimensionality counting to identify what the low-dimensional structure computes.

\section{Conclusion}
\label{sec:conclusion}

We introduced a functional perspective on training dynamics, showing that spectral edge directions correspond to low-dimensional modes in input space rather than localized structures in parameter or activation space.
This resolves a persistent mismatch in interpretability: while standard methods seek structure in representations, the dominant objects of learning may instead live at the level of functions.

Our experiments reveal a clear hierarchy.
When the task aligns with a group structure, functional modes collapse to single Fourier components in the appropriate basis.
When such alignment is absent, the learned structure remains low-dimensional but spreads across multiple modes or compositional subspaces.
Multitask training further demonstrates that these modes are reusable, supporting the view that neural networks build complex computations by composing simpler functional primitives.

This perspective suggests several directions for future work.
First, extending the analysis to language models may reveal whether similar functional subspaces underlie natural language processing.
Second, identifying the symmetry groups or algebraic structures of more complex tasks could provide predictive insight into the structure of learned representations.
Third, a systematic multitask ablation---training $x^2+y^2$ with only addition, only multiplication, or both---would isolate which subcircuit donates which functional mode.
Finally, developing tools that directly operate in function space may bridge the gap between training dynamics and interpretability.

More broadly, our results indicate that understanding neural networks may require moving beyond representation-level descriptions toward a theory of functional modes of learning.


\bibliographystyle{plainnat}

\appendix

\section{Co-Usage Analysis Details}
\label{sec:app_cousage}

For each test input $x = (a,b)$ and spectral direction $v_k$, we define the per-example activation $a_k(x) = \langle \nabla_\theta \ell(x),\, v_k \rangle$.
The \emph{co-usage matrix} $M_{ij} = \mathrm{corr}(a_i, a_j)$ captures whether two directions are activated by the same inputs.

In Olah et~al.'s framework \citep{olah2025interference}, the raw singular value $\sigma_k$ is analogous to a ``virtual weight''---its magnitude in parameter space.
The coactivation-corrected importance $\sigma_k \times d_k$ (where $d_k$ measures diagonal dominance of the co-usage matrix in the neighborhood of $v_k$) is analogous to the ``effective weight'' after accounting for interference.

In multitask models, we additionally compute the \emph{task-conditioned usage} $U_{o,k} = E_{x \sim \text{task } o}[a_k(x)^2]$ and the \emph{sharedness score}
\[
    S_k \;=\; \frac{\bigl(\sum_o U_{o,k}\bigr)^2}{T \cdot \sum_o U_{o,k}^2},
\]
ranging from $1/T$ (task-specific) to $1$ (equally shared across $T$ tasks).
Above-edge directions show high sharedness ($S \approx 0.8$--$0.9$); below-edge directions are more task-specific ($S \approx 0.3$--$0.5$).

We deliberately avoid the term ``circuits'' for spectral edge directions.
Section~\ref{sec:mech_fail} established that these directions are not localized in any standard structural sense.
The distinction is between functionally resolved and functionally unresolved modes, not between identified and unidentified circuits.

\end{document}